\documentclass[9pt,twocolumn,twoside]{pnas-new}

\usepackage[absolute]{textpos}

\definecolor{pink}{cmyk}{0, 1, 0, 0} 

\newcommand{\R}{\mathbb{R}}
\newcommand{\C}{\mathbb{C}}
\newcommand{\E}{\mathbb{E}}
\newcommand{\p}{\mathbb{P}}
\newcommand{\1}{\mathbf{1}}
\newcommand{\cN}{\mathcal{N}}
\newcommand{\ud}{\mathrm{d}}

\templatetype{pnasresearcharticle} 

\title{Gaussian Determinantal Processes: a new model for directionality in data}

\author[a,1]{Subhroshekhar Ghosh}
\author[b]{Philippe Rigollet} 

\affil[a]{National University of Singapore,
Department of Mathematics,
10 Lower Kent Ridge Road,
Singapore 119076
}
\affil[b]{Massachusetts Institute of Technology,
Department of Mathematics,
77 Massachusetts Avenue,
Cambridge, MA, 02139}

\leadauthor{Ghosh} 

\significancestatement{The increasingly complex nature of data has led statisticians to rethinking even the most basic of modeling assumptions. In this context, a determinantal point process (DPP) modeling paradigm promotes diversity in the sample at hand. In this work, 
we introduce a simple and flexible Gaussian DPP model to capture directionality in the data. Using the Gaussian DPP as an ansatz, we obtain a novel approach for dimensionality reduction that produces a better and more readable representation of the original data than standard Principal Component Analysis (PCA). These findings are supported by a finite sample analysis of the performance of our estimator, in particular in a spiked model similar to the one employed to analyze PCA.}

\authorcontributions{S.G. and P.R. designed research, performed research, analyzed data, and wrote the paper.}
\authordeclaration{The authors declare no conflict of interest.}
\correspondingauthor{\textsuperscript{1}To whom correspondence should be addressed. E-mail: subhrowork{@}gmail.com}


\begin{abstract}
Determinantal point processes (a.k.a. DPPs)  have recently become popular tools for modeling the phenomenon of negative dependence, or repulsion, in data. However, our understanding of an analogue of a classical parametric statistical theory is rather limited for this class of models. 
In this work, we investigate a parametric family of Gaussian DPPs with a clearly interpretable effect of parametric modulation on the observed points. We show that parameter modulation impacts the observed points by introducing directionality in their repulsion structure, and the principal directions  correspond to the directions of maximal (i.e. the most long ranged) dependency. 

\vspace{4pt}

\hspace{10pt}
This model readily yields a novel and viable alternative to Principal Component Analysis (PCA) as a dimension reduction tool that favors directions along which the data is most spread out. This methodological contribution is complemented by a statistical analysis of a spiked model similar to that employed for covariance matrices as a framework to study PCA.
These theoretical investigations unveil intriguing questions for further examination in random matrix theory, stochastic geometry and related topics.     
\end{abstract}

\doi{\href{https://doi.org/10.1073/pnas.1917151117}{www.pnas.org/cgi/doi/10.1073/pnas.1917151117}}

\begin{document}

\maketitle
\thispagestyle{firststyle}
\ifthenelse{\boolean{shortarticle}}{\ifthenelse{\boolean{singlecolumn}}{\abscontentformatted}{\abscontent}}{}


\begin{textblock}{10}(7.8,0.5) For published version see \href{https://doi.org/10.1073/pnas.1917151117}{\bf PNAS 117, no. 24 (2020): 13207-13213.}
\end{textblock}

\dropcap{T}he easy access to vast computational, data collection, and storage resources has fueled a paradigm shift in our approach to data-driven decision-making,  requiring us to rethink many modeling assumptions that are at the core of standard statistical techniques. In this paper, we investigate a certain class of determinantal point processes (DPPs) as an effective paradigm in understanding and analyzing directional effects in large data sets.

Determinantal processes have emerged as an important class of models in the recent years in probability, statistical physics and applied mathematics. Originally introduced to model Fermionic particle systems in physics, DPPs are powerful tools to capture repulsive interactions between the components of a system. A DPP  is a random set of points  characterized by the fact that its $k$-point correlation functions $\rho_k(x_1,\cdots,x_k)$ (roughly, the probability densities of having random points at $x_1,\cdots,x_k$) are given by determinants $\det[(K(x_i,x_j))_{1\le i,j\le k}]$ for some kernel $K$. A more detailed account of these processes is given in next section.

In recent years, DPPs have been examined as an effective tool to model certain phenomena like data diversity in statistics~\cite{LavMolRub15} and machine learning~\cite{KulTas12, AffFox14, Gil14, MarSra15, AffFox13}. DPPs are also closely associated with Random Matrix Theory (RMT) models and Coulomb systems, which turn out to be effective models in many practical scenarios that exhibit repulsive behavior between agents. Real world applications of RMT and Coulomb models are as varied as the study of transportation systems (in particular, the famous bus system of Cuernavaca), parking of cars, study of pedestrians, perched birds, geography, genetics, interacting particle systems, numerical computations, nuclear spectra and the theory of dynamical games - for detailed discussions we refer the reader to \cite{KrbSeb00, BaiBor06, JagTro17, Abu06, Jez10, Seb09, LecaDe93, LuZh06, Dei08, Dei17, W67, Ede13, vNGo47, DeiMen14, CarCer18} and the references therein. However, the  application of sophisticated models as DPPs  to address many standard statistical questions, such as \emph{dimension reduction}, remains a challenge and very much a work in progress. For instance, the assumption of independent observations has been essential in the development and analysis of many crucial statistical tools based on the concentration of measure phenomenon~\cite{Led01} on the one hand, as well as the establishment of fundamental limitations of statistical methods using the minimax paradigm~\cite{Tsy09}. Our understanding of such techniques and their applications for models like DPPs are much more limited in comparison.

In this paper, we propose a new family of determinantal probability measures, called Gaussian Determinantal Processes, on $\mathbb{R}^d$ that is parametrized by a so-called \emph{scattering matrix} of size $d\times d$. In particular, this positive-definite matrix captures the directionality of the DPP in the following sense: its eigenvectors associated to its largest eigenvalues span low-dimensional spaces along which the DPP exhibits the most repulsion. We propose a consistent estimator of the scattering matrix and demonstrate that these eigenspaces produce low-dimensional representation of the original data that are, in some sense, better than Principal Component Analysis (PCA) on two benchmark data-sets: Fisher's Iris data and the Wisconsin Breast Cancer data set. In this  context, we employ the DPP model merely as an \emph{ansatz} to serve as a basis for the development of a new method. This similar in spirit to the celebrated Wigner surmise  \cite{Wigner55,W67,mehta2004random}, wherein the model of Gaussian random matrices was introduced to approximate a specific aspect of heavy nuclei (namely, the statistics of their spectral gaps), rather than purporting to model the actual nuclear Hamiltonians themselves. Another analogy would be classical PCA itself, wherein the fundamental concepts are motivated and inspired by  Normal random variables, but are effectively employed in settings far beyond its Gaussian origins.  

These findings are supported by a statistical analysis of Gaussian determinantal processes in a \emph{spiked} model that is similar to one employed to study PCA and that has been the theatre for important developments in random matrix theory such as the celebrated BBP phase transition~\cite{BaiBenPec05}. In particular, despite the lack of independence in the data generated by a DPP, we show that, akin to the spiked covariance model for traditional PCA, the spike in this DPP model may be detected at at a signal strength proportional to $1/\sqrt{n}$---up to logarithmic factors---where $n$ is the number of observations in the DPP model.

\section*{Gaussian determinantal processes}

In this work, we are concerned with DPPs on $\R^d$. Recall that, when it exists, the family of  $k$-point correlation functions $\rho_k(x_1,\cdots,x_k), k \ge 1$, characterizes the distribution of point process $X$ as follows. For any integer $k\ge 1$ and any set of $k$ disjoint Borel subsets $A_1, \ldots, A_k$ of $\R^d$, let $X(A_i)$ denote the number of points of $X$ that are included in $A_i$. Then
$$
\E[X(A_1)\cdots X(A_k)]=\int_{A_1\times \cdot \times A_k}\rho_k(x_1,\cdots,x_k)\ud x_1 \cdots \ud x_k\,.
$$
A point process $X$ on $\R^d$ is called a DPP if there exits a kernel $K:\R^d\times \R^d\to \C$ such that $X$ admits $k$-point correlations $\rho_k$ given by
$$
\rho_k(x_1,\cdots,x_k)=\det\left[\begin{array}{ccc}K(x_1,x_1) & \cdots & K(x_1, x_k) \\\vdots & \ddots & \vdots \\K(x_k,x_1) & \cdots & K(x_k,x_k)\end{array}\right]\,.
$$
For more on the theory of point processes in general, and DPPs in particular, we direct the interested reader to~\cite{HouKriPer09, Kallenberg, Bor11}.

We are now in a position to define Gaussian determinantal processes. To that end, let $\Phi$ denote the density of a multivariate Gaussian distribution with mean 0 and covariance matrix $\Sigma$:
$$
\Phi(u)=\frac{1}{(2\pi)^{\frac{d}{2}} \sqrt{\det{\Sigma}}}\exp\big(-\frac{1}{2}u^\top \Sigma^{-1}u\big)\,, \quad u \in \R^d\,.
$$
We say that the point process $X$ over $\R^d$ is a \emph{Gaussian determinantal processes (GDP)} if it is a DPP with kernel $K(x,y)=\Phi(x-y)$ for some positive definite matrix $\Sigma$, called the \emph{scattering matrix} of $X$. 

Note that the density, i.e., the average number of points per unit volume, is given by $K(x,x)=\Phi(0)$.
Throughout this paper, we assume the normalization 
\begin{equation}
\label{EQ:norm}
K(x,x)=\Phi(0)=\frac{1}{(2\pi)^{\frac{d}{2}} \sqrt{\det{\Sigma}}}=1\,,
\end{equation}
in order to focus on the spatial dependence structure of the points---as opposed to the spatial scale that is captured by the density. This corresponds to a simple rescaling of the coordinate axes by the same constant factor and, importantly, does not affect the eigenvectors of $\Sigma$, which are used for dimension reduction. Note that other normalizations may be considered by simply multiplying $\Phi$ with an appropriate constant. This was done, for example, in~\cite{LavMolRub15} who introduced GDPs in the isotropic case where $\Sigma=I_d$ and therefore did not focus on directionality as is the case here, but rather on density as a variable parameter. Using Fourier analytic techniques to understand the spectrum of $K$ (viewed as an integral operator) together with the celebrated Macchi-Soshnikov Theorem~\cite[Theorem 4.5.5]{HouKriPer09}, it can be shown GDPs exist for any positive definite scattering matrix (See Theorem~1 in the supporting information).  By construction, GDPs are stationary point processes, that is, their distribution is invariant under translations. 

The goal of Gaussian DPP is to capture the spatial dependency structure of data via the scattering matrix $\Sigma$. To see how this manifests itself, observe first that when $\Sigma=\sigma^2 I_d$ for some $\sigma>0$, we have $K(Ux,Uy)=K(x,y)$ for every orthogonal matrix $U$, and hence that the distribution of this DPP is also invariant under rotations. To understand the effect of $\Sigma$ on directionality at a qualitative level, let us consider the simplest measure of dependence of the data: the two point correlation function $\rho_2$. As we already noted, this is roughly the probability density of having a point at $x$ and a point  at $y$. It is customary in statistical physics to pass from $\rho_2$ to the truncated pair correlation $\bar \rho_2$, given by $\bar \rho_2(x,y)=\rho_2(x,y)-\rho_1(x)\rho_1(y)$. This amounts to subtracting off the independent part of the pair correlation, which is roughly the probability  of having  points at $x$ and $y$ if there was no spatial dependence, and this contribution is simply $\rho_1(x)\rho_1(y)$. Thus $\bar \rho_2$ captures the pure contribution of the dependence structure between the points. For a determinantal process with kernel $K$, it is an immediate observation that $\bar \rho_2=-|K(x,y)|^2$. This in particular means that for a DPP, it holds $\bar \rho_2\le 0$, which captures the spatial repulsion that is characteristic of such processes. The smaller $\bar \rho_2(x,y)$  is in magnitude, the more decorrelated (and hence Poissonian) the point field at $x$ and $y$ is. Conversely, the bigger $\bar \rho_2(x,y)$ is in magnitude, the more strongly correlated is the process. 

For GDPs (with the normalization [~\eqref{EQ:norm}]), the truncated pair correlation is given by
$$
\bar \rho_2(x,y)= - \exp\big(- (x-y)^T \Sigma^{-1} (x-y)  \big)\,. 
$$
Therefore, the magnitude of $\bar \rho_2(x,y)$ is proportional to $\phi_{\frac{1}{2}\Sigma}(x-y)$ where $\phi_{\frac{1}{2}\Sigma}$ is the density of a centered multivariate Gaussian distribution with covariance matrix $\frac{1}{2} \cdot \Sigma$. In particular, if $x-y$ is well aligned with an eigenvector of $\Sigma$ associated to a relatively large eigenvalue, then the magnitude of $\bar \rho_2(x,y)$ is large even for large values of $\|x-y\|$. In other words, for such directions, the spatial correlation has longer range.

This motivates us to formulate a spiked model to study the estimation of the spectrum of the scattering matrix $\Sigma$.
It follows from the above discussion that the eigenvectors of the scattering matrix $\Sigma$ that are associated to large eigenvalues capture long-range spatial correlations---more specifically repulsion---between the points generated by the GDP. This phenomenon is the basis of a \emph{spiked scattering model} by analogy to the popular spiked covariance model~\cite{Joh01,Pau07,FerPec09} where the covariance matrix $S$ is assumed to be a rank-one perturbation of the identity: $S=I_d + \lambda uu^\top$, $\|u\|=1, \lambda\ge 0$. In this case, all the directionality of the model is carried by the rank-one perturbation $uu^\top$ called the \emph{spike} and its strength is carried by the parameter $\lambda>0$.

In the context of GDPs, it is natural to work in the situation where the presence of the spike leaves the mean density of points unchanged compared to the isotropic case, where the scattering matrix  is the identity. If that is not the case, then the presence of the spike can be detected simply by estimating the density. The mean density remaining unchanged amounts to $\det(\Sigma)$ being equal for the null and alternatives.  This leads to the model 
\begin{equation}
\label{EQ:spike}
(2\pi)\Sigma=(1+\lambda)^{-\frac1{d-1}}(I_d-uu^\top) + (1+\lambda) uu^\top\,, \quad \|u\|=1\,,
\end{equation}
where $\lambda\ge 0$ is the strength parameter and $uu^\top$ is the spike. Notice that when $\lambda =0$, we recover the isotropic case $\Sigma=I_d$. The constant $2\pi$ in front of $\Sigma$ is inconsequential and simply ensures the normalization adopted [~\eqref{EQ:norm}].


Under the spiked model [~\eqref{EQ:spike}], we ask two natural statistical questions: 
\begin{itemize}
\item[(i)] The \emph{detection} question consists in studying the signal strength $\lambda>0$ sufficient to detect the presence of the spike from data (the precise data scheme is described in detail in the next section); this is a simple-vs-composite hypothesis testing problem of the form:
$$
\begin{array}{rl}
H_0:& (2\pi)\Sigma=I_d \qquad \text{vs.}\\
H_1:& (2\pi)\Sigma=(1+\lambda)^{-\frac1{d-1}}(I_d-uu^\top) + (1+\lambda) uu^\top\,, \\ & \quad \quad \quad \quad  \|u\|=1\,.
\end{array}
$$
To define this problem more precisely, let $\p_0$ (resp. $\p_u$) denote the probability distribution associated to the GDP with scattering matrix $I_d$ (resp. $\lambda^{-\frac1{d-1}}(I_d-uu^\top) + \lambda uu^\top$). We say that we can \emph{detect} the spike at strength $\lambda>0$ if for any $\delta \in (0,1)$, there exists a Borel function of the data $\psi\in \{0,1\}$, called \emph{test}, such that
\begin{equation}
\label{EQ:testerror}
\p_0(\psi=1)\vee \sup_{u:\|u\|=1}\p_u(\psi=0) \le \delta\,.
\end{equation}
\item[(ii)] The \emph{estimation} question consists in estimating the direction $u$ at a constant signal strength $\lambda$. Such results follow from a good estimator of $\Sigma$ together with matrix perturbation arguments.
\end{itemize}

Both questions tie in naturally with the spiked covariance matrix models that are nowadays ubiquitous in high dimensional statistics, particularly in the context of Principal Component Analysis or PCA~\cite{And84,BerRig13,BerRig13b,JolCad16}. 

\section*{Statistical estimation}

In spite of the clarity of the underlying principle behind DPPs, our understanding of fundamental statistical procedures such as the method of moments and maximum likelihood estimation in this context is largely limited to the discrete settings~\cite{BruMoiRig17,UrsBruMoi17} when \emph{multiple independent realizations} of the process are available to the statistician. While this setup is particularly relevant to machine learning, it leaves behind the fundamental framework of continuous DPPs where a single realization of the process is observed~\cite{LavMolRub15}. This framework poses a significant challenge due to the complete lack of independence between the sample points.

Note that by stationarity, a DPP realization has almost surely an infinite number of points over $\R^d$.  In this work, we assume that  we observe the  realization of this DPP in a Euclidean ball $B(R) \subset \R^d$ centered at the origin and with radius $R>0$.

Our estimation strategy is driven by the fact that we can examine many distinct local neighborhoods of our observation window. By translation-invariance, we may average various statistics over such local neighborhoods to recover global properties of the DPP.

Let $X$ be a GDP with scattering matrix $\Sigma$ and let $X_1,\ldots, X_{N} \in B(R)$ denote the set of points of $X$ that are included in $B(R)$. Note that $N$ is a random number but it concentrates sharply (Cf. Theorem~2 in SI) around its expectation 
$$
n:=\E[N]=|B(R)|=|B(1)|R^d\,,
$$
where $|B(t)|$ denotes the volume of a Euclidean ball with radius $t \ge 0$.

We are now in a position to define an estimator $\hat \Sigma$ for the scattering matrix $\Sigma$. It is parametrized by a positive cut-off threshold $r<R$ that is chosen theoretically by realizing a bias-variance tradeoff. However, 
empirical investigations indicate that our estimator shows little sensitivity to the choice of $r$.

For each observation $X_i, i=1, \ldots, N$, let $\cN_i$ denote the set of other observations that are at distance at most $r$ from it: 
$$
\cN_i=\{j\,:\,j\ne i,  \,\|X_i-X_j\|<r\}\,.
$$
To avoid boundary effects, we also define $\cN_0=\{j\,:\, \|X_j\|<R-r\}$.
Define the estimator $\hat \Sigma$ of $\Sigma$ to be the $d\times d$ matrix given by
$$
\hat \Sigma = |B(1)|\frac{r^{d+2}}{d+2} I_d-  \frac{1}{|B(R-r)|}\sum_{i \in \cN_0} \sum_{j \in \cN_i} (X_i-X_j)(X_i-X_j)^\top .
$$

Using  the Fourier analytic properties of the Gaussian kernel, we can control the accuracy of the  the estimator $\hat{\Sigma}$ in Frobenius norm. Throughout our considerations, we assume that $\|\Sigma\|_{\mathrm{op}}$ is bounded above by a universal constant. 
This amounts to assuming that all the eigenvalues of $\Sigma$ are of order one, or in other words, that the scattering matrix is not degenerate.

This estimator is negatively biased in the sense that for any unit vector $v \in \R^d$ we have $\E[v^\top\hat{\Sigma}v] \le v^\top {\Sigma}v^\top$. Moreover, it can be shown that, for $r$ large enough we have
$$
\|\E\hat{\Sigma}-\Sigma\|^2_{\mathrm{F}}\lesssim d \exp\big(C(2d-r^2))\big)\,, 
$$
for some positive constant $C$. In particular, the bias vanishes exponentially fast as $r \to \infty$. For details on these results, we refer the reader to see Lemma~4 in SI. 

Moreover, the variance of $\hat{\Sigma}$ can be controlled as follows (Lemma~5 in SI): For $r$ large enough,
$$
\E\|\hat{\Sigma} - \E\hat{\Sigma}\|_{\mathrm{F}}^2\le d^2\left(\frac{C}{d}\right)^d\frac{r^{2d+4}}{n}\,.
$$
In particular, the bound on the variance is inversely proportional to the sample size as is the case in the independent setting.

If we choose the $r=C\sqrt{d \log n}$ for a universal constant $C>0$ so as to realize the bias-variance tradeoff (Theorem~3 in SI), we get
\begin{equation}
\label{EQ:delta}
\E \| \hat{\Sigma} - \Sigma \|_{\mathrm{F}}  \le \mathfrak{r}_{n,d}:=\frac{d^2(c \sqrt{\log n})^{d+1}}{\sqrt{n}}, \quad c>0\,,
\end{equation}
for $n$ large enough. As a result, we actually get a non-asymptotic bound $\E \| \hat{\Sigma} - \Sigma \|_{\mathrm{F}}$ that is of order $1/\sqrt{n}$ up to logarithmic terms as in the traditional i.i.d setting, despite the dependence between points.
However, the (exponential) dependence on the dimension of the bound in this result leaves scope for improvement in further investigations.

The estimator $\hat{\Sigma}$ might fail to be positive definite, albeit with a small probability that vanishes as the data size tends to infinity, because of the concentration of $\hat{\Sigma}$ around $\Sigma$. If, for some application, it is necessary to have a positive definite estimator, one can naturally consider the projection of $\hat{\Sigma}$ on to the positive definite cone.

\section*{The spiked model}

We now turn to the spiked model~\eqref{EQ:spike}. 

\subsection*{Detection}

It is useful to note that in the hypothesis testing associated to detection, it holds that $\|\Sigma\|_{\mathrm{op}}=1/2\pi$ under $H_0$ and $\|\Sigma\|_{\mathrm{op}}=(1+\lambda)/2\pi$ under $H_1$. Of course, the above result readily yields that $\E\| \hat{\Sigma} - \Sigma \|_{\mathrm{op}} \le  C(d)/\sqrt{n}$. This leads us to consider the test statistic $\|\hat \Sigma\|_{\mathrm{op}}$ and the test
$$
\psi_t=\1\big\{ (2\pi)\|\hat \Sigma\|_{\mathrm{op}}>1+t\mathfrak{r}_{n,d}\big\}\,, \quad t>0\, ,
$$
where $\1\{\cdot\}$ is the indicator function.
Using the Markov and triangle inequalities, we get that if $t=1/\delta$ and 
$$
\lambda>\frac{\mathfrak{r}_{n,d}}{\pi}\,,
$$
then the text $\psi_{1/\delta}$ satisfies~\eqref{EQ:testerror}. In particular, this implies that we can detect the presence of the spike at strength $\bar\lambda=2\mathfrak{r}_{n,d}$. 

Our technique,  based on bounding the operator norm by the Frobenius norm, is too blunt to characterize the sharp dependence in the dimension $d$ of the optimal strength at which the spike may be detected. An answer to this question would involve a sophisticated understanding of the large deviations of $\hat \Sigma$, requiring detailed investigations in its own right. We believe that this would lead to challenging problems in random matrix theory, involving the understanding of the strong spatial dependence and the stochastic geometry of GDPs.

\subsection*{Estimation}

When $\lambda$ is large enough, the estimation of $u$ in the model~\eqref{EQ:spike} follows from matrix perturbation analysis. Recall that $u$ is the eigenvector associated to the largest eigenvalue $1+\lambda$ of $\Sigma$ and that $\Sigma$ exhibits a spectral gap of $1+\lambda-(1+\lambda)^{-\frac{1}{d+1}}\ge \lambda$. Moreover, let $\hat u$ denote any eigenvector associated to the largest eigenvalue of $\hat \Sigma$. it follows from the Davis-Kahan theorem that
$$
\E|\sin(\measuredangle(\hat u, u))| \le \frac{\E\| \hat{\Sigma} - \Sigma \|_{\mathrm{op}}}{\lambda}\le \frac{\mathfrak{r}_{n,d}}{\lambda}\,,
$$
where $\measuredangle(\hat u, u)$ denotes the angle between $\hat u$ and $u$. In light of~\eqref{EQ:delta}, the above result indicates that we can estimate the spike at the parametric rate $1/\sqrt{n}$. As before, the optimal dependence in the dimension $d$ is left for future research. Nevertheless, the above result gives us a theoretical basis for dimension reduction technique presented in the next section.

\section*{Dimension reduction}

To illustrate the ability of the GDP model to find directionality in data, we apply our techniques to two benchmark data sets: Fisher's Iris data and the Wisconsin Breast Cancer data set. We compare our results with the representations obtained by standard PCA.

\begin{figure}[h]
\centering
\includegraphics[width=.48\linewidth]{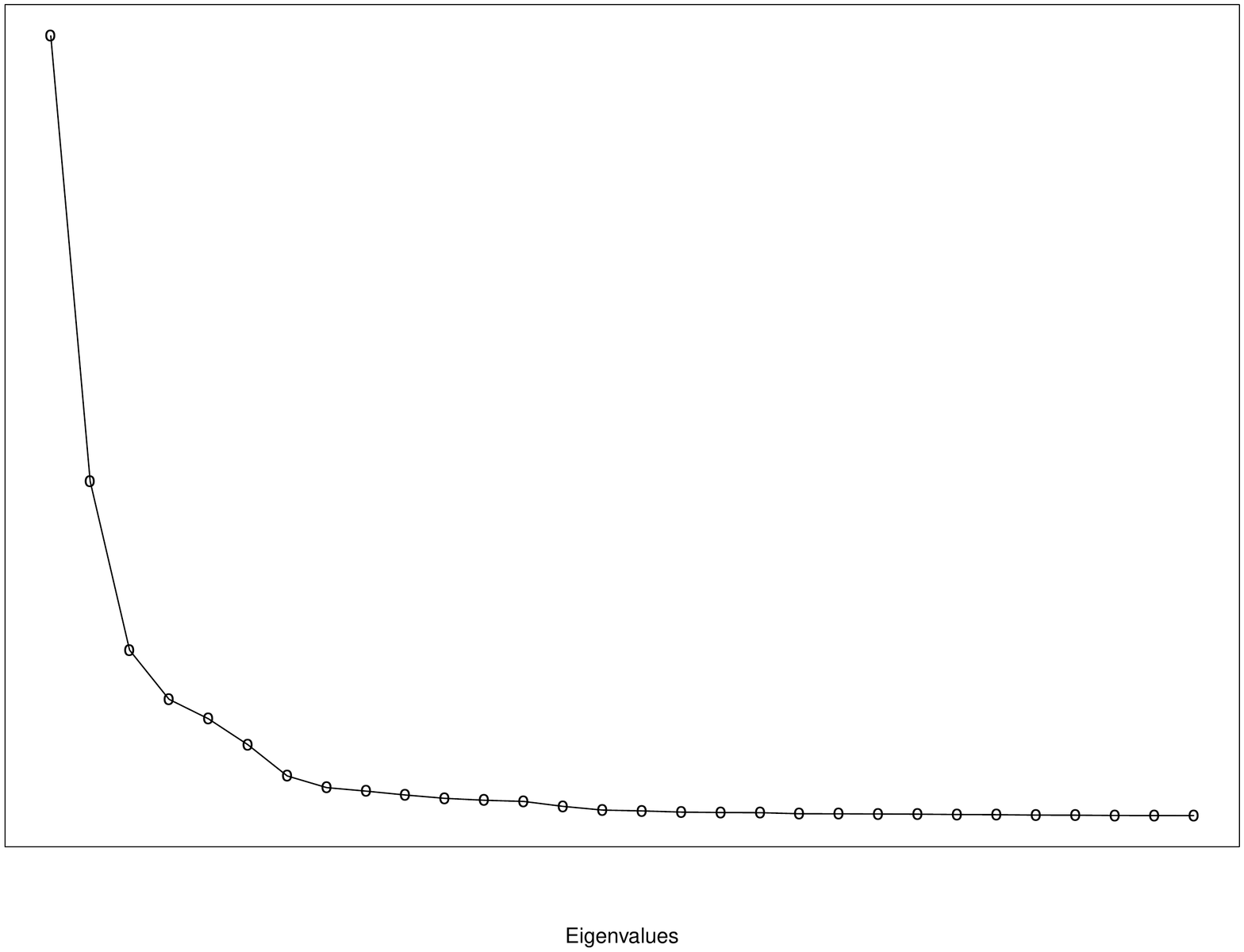}\ \includegraphics[width=.48\linewidth]{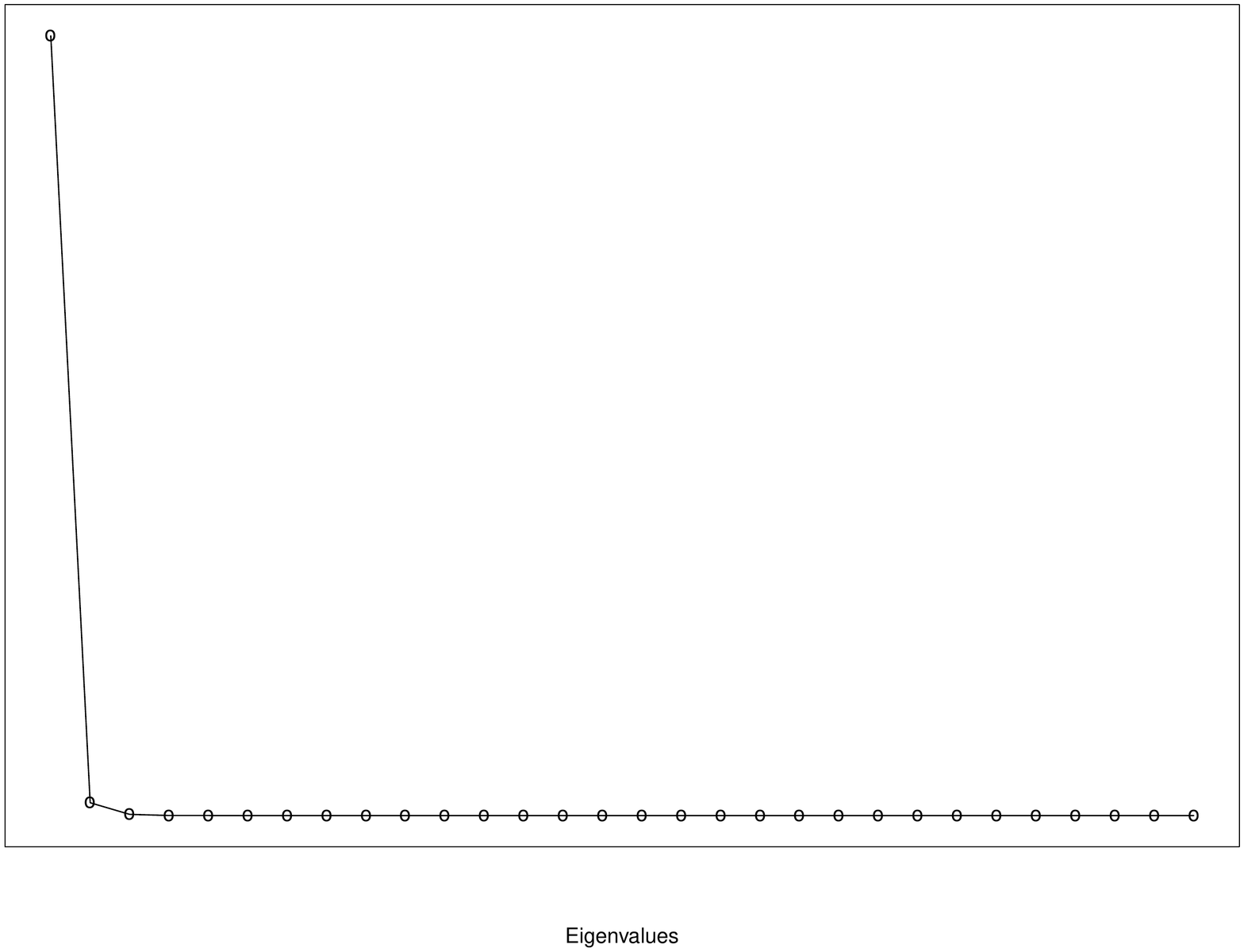}
\caption{Ordered eigenvalues of the sample correlation matrix (left) and of  $\hat \Sigma$ (right)}
\label{FIG:scree}
\end{figure}

More precisely, given the data, we estimate the matrix $\Sigma$ using the estimator $\hat \Sigma$ defined above. Note that for the purpose of computing eigenvectors of $\hat \Sigma$, the choice of $R$ does not matter. Moreover, avoid any parameter tuning, we take $r$ large enough so that all the points are in the neighborhood $\cN_i$ of all other points $i$. While this choice minimizes the bias term, it potentially allows for a large variance. However, this choice gave the best result in both numerical examples below. It appears from the singular value decomposition of our estimator that the non-degeneracy assumption that we make in our analysis is violated, leading to a large operator norm for $\Sigma$ (see Figure~\ref{FIG:scree}). In particular, this contributes to the upper bound on the bias and can be compensated by a large value of $r$, thus explaining the good empirical behavior of a large $r$.

Then we compute the singular value decomposition of $\hat \Sigma$ and extract the eigenvectors associated to the largest eigenvalues. In both case, it is worth noting that the leading eigenvalue far exceeds the subsequent ones, which indicates that for these two datasets, most of the repulsion is captured by one dimension (see Figure~\ref{FIG:scree}).

For both datasets, when displaying a 2D scatterplot, we simply project the original points on the space spanned by the first two eigenvectors of $\hat \Sigma$ and compare with the representation given by PCA (with centering and scaling options activated).

\subsection*{Fisher's Iris}

\begin{figure}[h!]
\centering
\includegraphics[width=.48\linewidth]{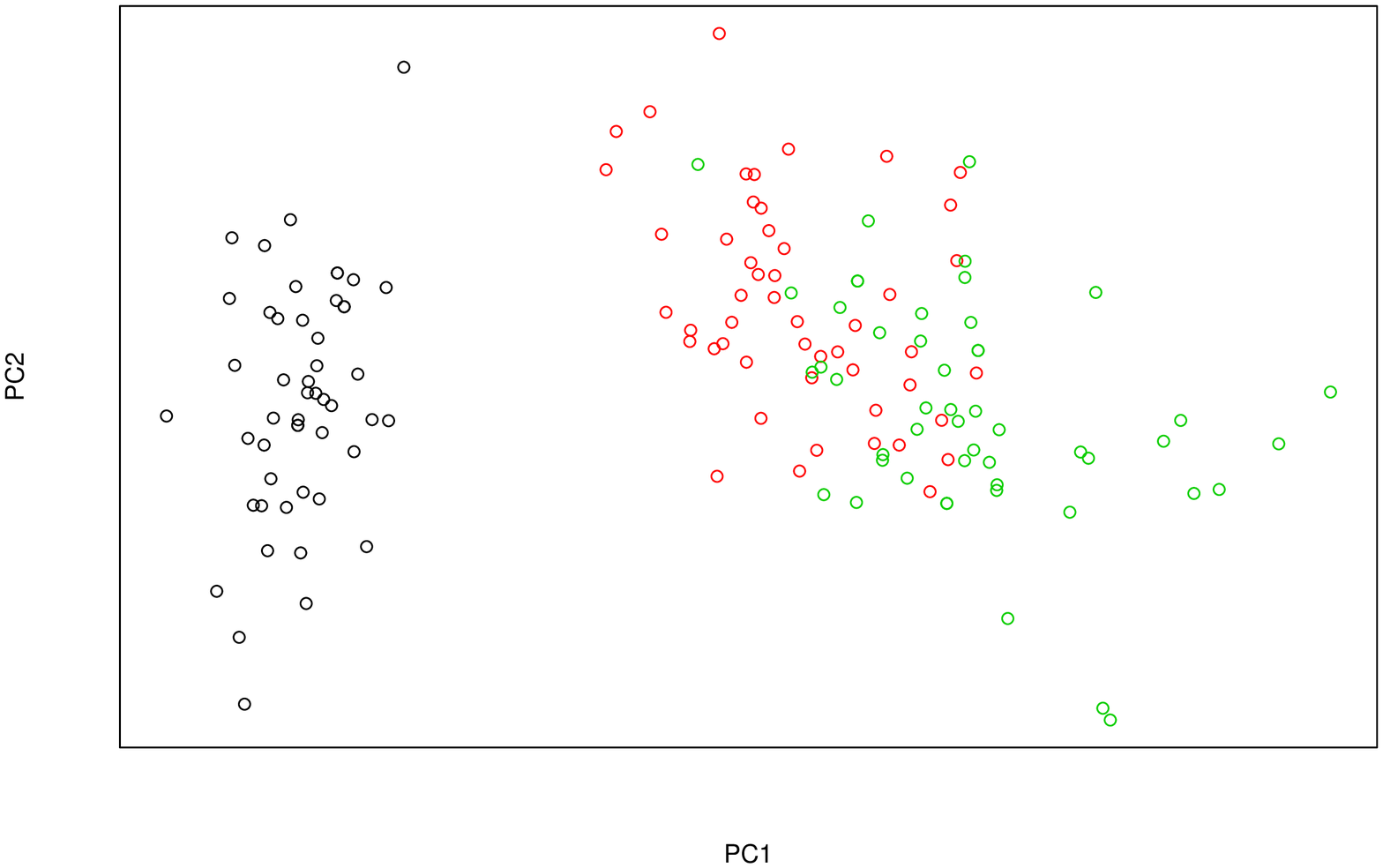}\ \includegraphics[width=.48\linewidth]{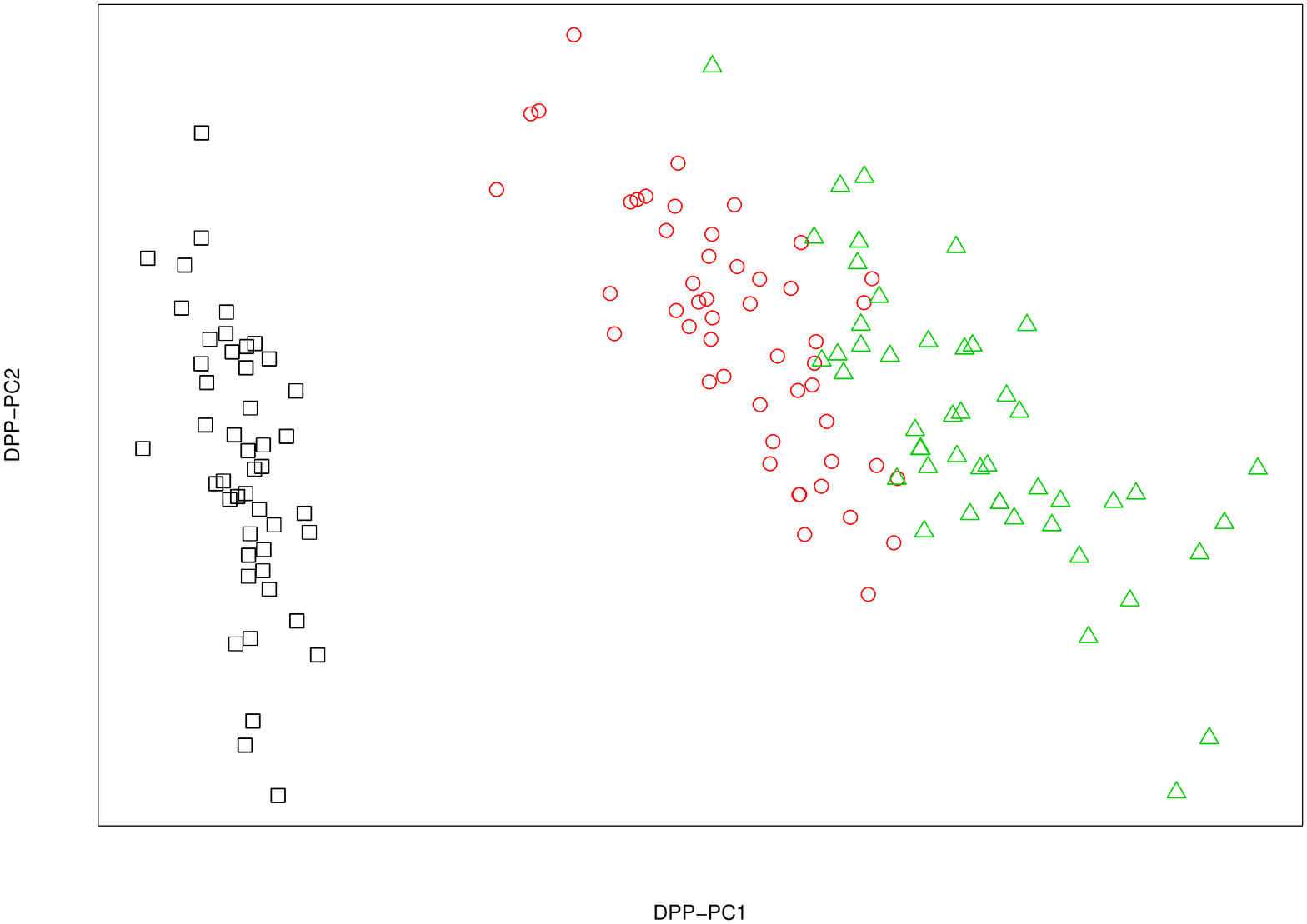}
\caption{Projection of the first two principal components (left) and the first two eigenvectors of $\hat \Sigma$ (right)}
\label{FIG:iris}
\end{figure}

While this dataset contains little mystery, it is perhaps the most standard data set to apply PCA to ~\cite{FIdata, Fis36}. It contains $N=150$ observations in dimension $d=3$ split into three clusters, each corresponding to a different type of iris. In Figure~\ref{FIG:iris} the points are colored according the the cluster they belong to. While the main structure between the clusters is preserved between the two representations, two phenomena emerge. On the one hand, the DPP approach leads to a point-cloud that displays better spatial organization, especially locally, due to the repulsion between the points in these directions. On the other hand, the DPP approach appears to provide a better clustering between the red circles (Versicolor) and the green triangles (Virginica).

\subsection*{Wisconsin Breast Cancer}
The Wisconsin Breast Cancer data set ~\cite{WBdata, Wol90} comprises of 569 observations in dimension $d=30$. Each observation corresponds to either a benign or a malignant tumor. This data set had originally been obtained from the University of Wisconsin Hospitals, Madison from Dr. William H. Wolberg. 

\begin{figure}[t]
\centering
\includegraphics[width=.48\linewidth]{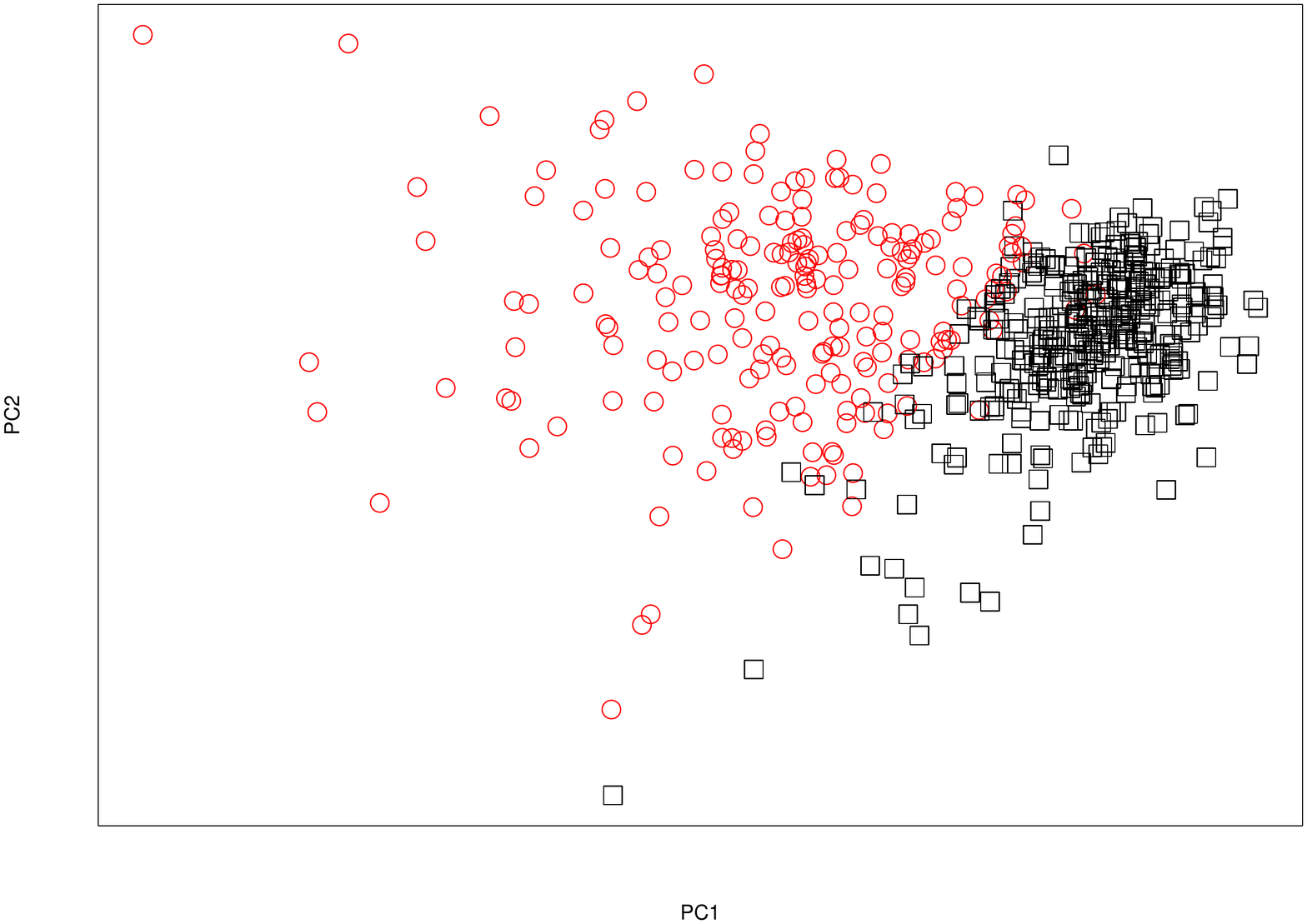}\ \includegraphics[width=.48\linewidth]{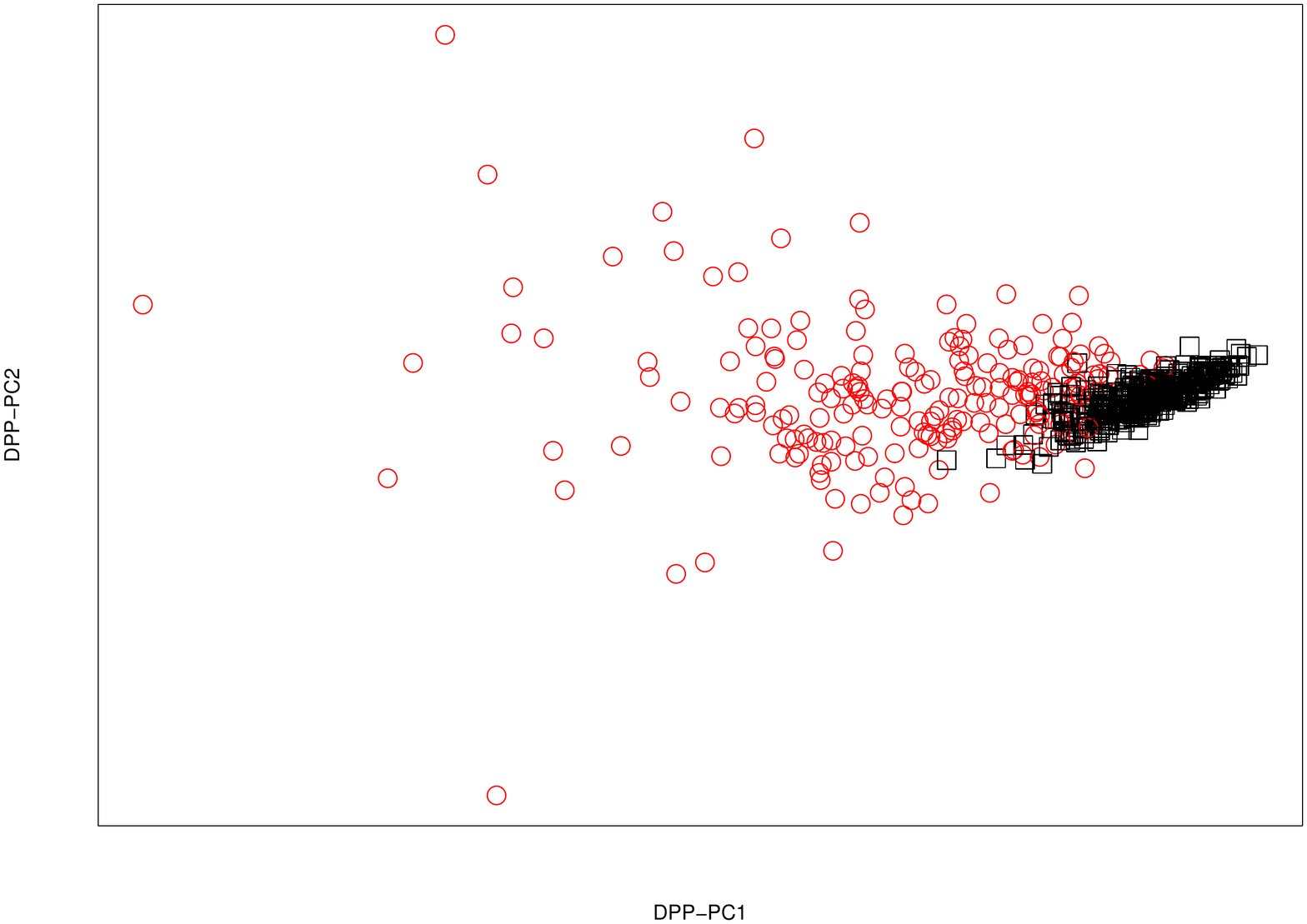}
\caption{Projection of the first two principal components (left) and the first two eigenvectors of $\hat \Sigma$ (right)}
\label{FIG:W}
\end{figure}

On this dataset, the difference between the two methods is stark as illustrated in Figure~\ref{FIG:W}. While it appears that the DPP-based approach presents less repulsion between the points, especially within the cluster of black squares, this should be mitigated with the fact that this cluster corresponds, in fact, to benign tumors. As a result, the DPP approach tends to cluster the benign tumors into one tight cluster whereas it maintains significant diversity within the cluster of malignant tumors. This phenomenon is not without resemblance to the opening line of Tolstoy's Anna Karenina: ``Happy families are all alike; every unhappy family is unhappy in its own way.''

\begin{figure}[h]
\centering
\includegraphics[width=.48\linewidth]{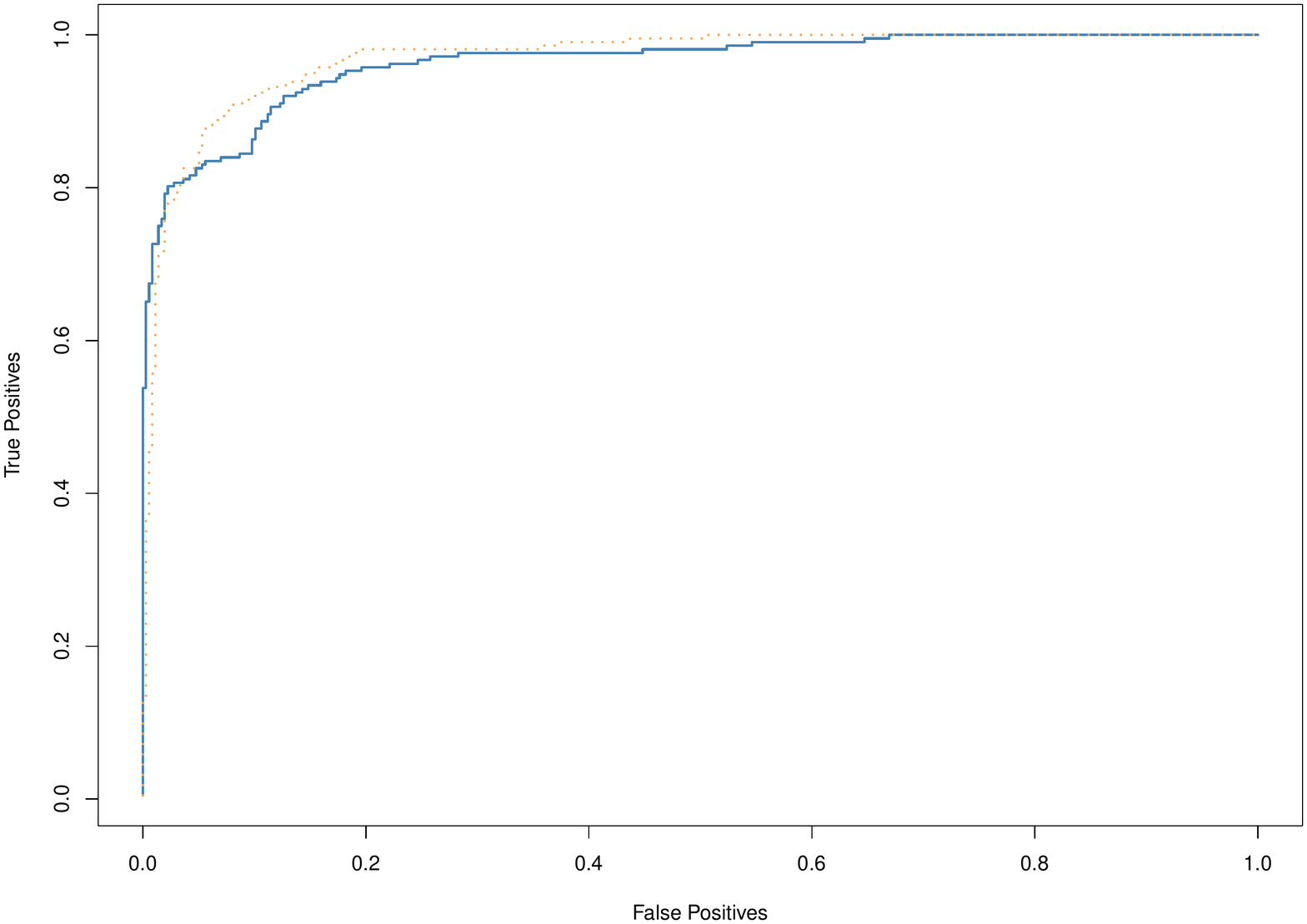}\ \includegraphics[width=.48\linewidth]{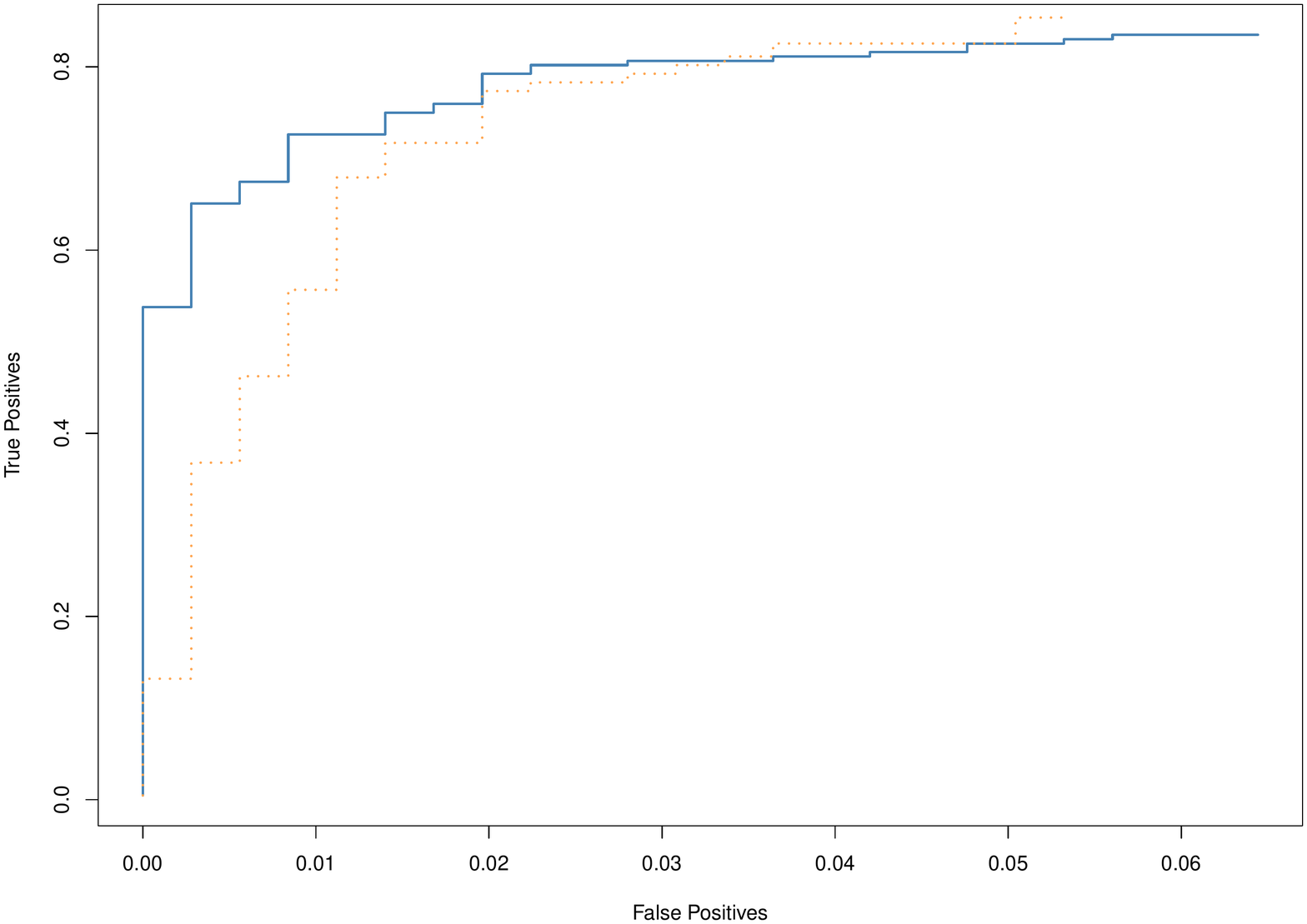}
\caption{Left: ROC curves given by the first principal component (dashed) and the first eigenvector of $\hat \Sigma$ from the DPP approach (solid). Right: only the 200 observations with the highest risk score are represented.}
\label{FIG:ROC}
\end{figure}

It turns out that the DPP approach also does a better job at separating benign from malignant tumors in the following sense. Consider the \emph{risk score} associated to an observation as the negative value of its coordinate when projected onto either the the first principal component or the leading eigenvector of $\hat \Sigma$. The larger the (negative) value for this coordinate, the higher the risk of the tumor being malignant. Receiver operating characteristic or ROC curves are a well established tool to compare risk scores. In Figure~\ref{FIG:ROC}, we compare the ROC curves obtained using the coordinates from the first principal component (PCA approach) and the coordinates from the leading eigenvector of $\hat \Sigma$ (DPP approach). These are comparable and both quite good:  the PCA approach gives an area under the ROC curve of .970 whereas the DPP approach gives one of .963. However, when zooming towards 
high scores (that is, greater risk of being malignant), we see in Figure~\ref{FIG:ROC} (right) that the DPP approach dominates the PCA, even displaying more than 50\% of true positives while having no false positives. This performance is all the more impressive that it is a fully unsupervised method.

For this dataset, it is also interesting to look at the associated scree plots which indicate the values of the eigenvalues from largest to smallest. While in the PCA case, we observe a graceful decay, the DPP approach outputs leading eigenvalue that largely dominates all others. This indicates that for this dataset, the spiked model~\ref{EQ:spike} appears to hold approximately.

\section*{Conclusion}
In this work, we investigated the Gaussian DPP model to capture directionality in the dependence structure of data. This parametric model is driven by a kernel that is parametrized by a $d \times d$ positive-definite scattering matrix, whose spectral properties govern the directionality inherent in the point set. We propose a consistent estimator of the scattering matrix, and employ the leading eigenvectors of this estimating matrix to obtain a low-dimensional representation of the data. Applying this methodology to two benchmark data sets, namely the Fisher's Iris data and the Wisconsin Breast Cancer data, we exhibit that the subspaces obtained from GDP show arguably better performance than the PCA as a dimension reduction technique. Our empirical investigations are complemented by a theoretical study of an accompanying spiked model for the scattering matrix, wherein we provide detection guarantees that hold with high probability (with increasing sample size $n$) at a detection threshold of $1/\sqrt{n}$ (upto logarithmic factors).  In spite of the significant challenges posed by the complete absence of any kind of independence in the DPP model, this is comparable to the analogous threshold for spiked covariance matrices.

The GDP model and its application as a tool for dimension reduction throws up natural questions and exciting new challenges for further investigation.
A natural problem is to extend the variance estimates for our estimator to CLTs and tight non-asymptotic bounds, so that a finer theoretical analysis of the estimation and testing problems can be carried out. From the modeling perspective, it would be natural to examine more complex alternatives for the spiked model compared to the fundamental 1-parameter case studied in this paper, including a richer interpolating family between the identity and the spike and the possibility of having multiple spike directions. Another direction to study would be the extension of the ideas contained in the present paper to more general kernel classes beyond the Gaussian. The investigation of the estimator $\hat{\Sigma}$ as a random matrix, in particular its spectral properties and their large deviations, would be of great theoretical interest in its own right, as well as have important implications for the statistical problems discussed in this work. These can be related to the general objective of improving of the dimensional dependence in the rates obtained in the present paper. Finally, in the context of the spiked model, establishing a sharp threshold for detection poses an natural question to investigate. We envisage that these investigations will lead to intriguing challenges in random matrix theory, stochastic geometry and the analysis of directionality in data, ultimately involving the tension between the lack of independence on one hand and the concrete determinantal structure of the GDP on the other.

\section*{Materials and Methods}
The data analyses discussed in this paper have been implemented in R. The relevant codes are publicly available at \cite{Rig19}. The two benchmark datasets employed in this paper are available at \cite{FIdata} and \cite{WBdata} for Fisher's Iris data and Wisconsin Breast Cancer data  respectively.

\acknow{S.G. is supported in part by the MOE grant R-146-000-250-133. P.R. is supported by NSF awards IIS-BIGDATA-1838071, DMS-1712596 and CCF-TRIPODS-1740751; ONR grant N00014-17- 1-2147. We thank Victor-Emmanuel Brunel for illuminating discussions. We thank the referees for their careful reading of the paper, and for their insightful comments and suggestions.}
\showacknow{} 

\bibliography{dpp}

\begin{thebibliography}{10}

\bibitem{LavMolRub15}
Lavancier F, M{\o}ller J, Rubak E (2015) Determinantal point process models and
  statistical inference.
\newblock {\em J. R. Stat. Soc. Ser. B. Stat. Methodol.} 77(4):853--877.

\bibitem{KulTas12}
Kulesza A, Taskar B (2012) {\em Determinantal Point Processes for Machine
  Learning}, Foundations and Trends in Machine Learning.
\newblock (Now Publishers Inc.) Vol.{}~5.

\bibitem{AffFox14}
Affandi RH, Fox E, Adams R, Taskar B (2014) Learning the parameters of
  determinantal point process kernels in {\em International Conference on
  Machine Learning}.
\newblock pp. 1224--1232.

\bibitem{Gil14}
Gillenwater JA (2014) Approximate inference for determinantal point processes.

\bibitem{MarSra15}
Mariet Z, Sra S (2015) Fixed-point algorithms for learning determinantal point
  processes in {\em International Conference on Machine Learning}.
\newblock pp. 2389--2397.

\bibitem{AffFox13}
Affandi RH, Fox E, Taskar B (2013) Approximate inference in continuous
  determinantal processes in {\em Advances in Neural Information Processing
  Systems}.
\newblock pp. 1430--1438.

\bibitem{KrbSeb00}
Krb{\'a}lek M, Seba P (2000) The statistical properties of the city transport
  in cuernavaca (mexico) and random matrix ensembles.
\newblock {\em Journal of Physics A: Mathematical and General} 33(26):L229.

\bibitem{BaiBor06}
Baik J, Borodin A, Deift P, Suidan T (2006) A model for the bus system in
  cuernavaca (mexico).
\newblock {\em Journal of Physics A: Mathematical and General} 39(28):8965.

\bibitem{JagTro17}
Jagannath A, Trogdon T (2017) Random matrices and the new york city subway
  system.
\newblock {\em Physical Review E} 96(3):030101.

\bibitem{Abu06}
Abul-Magd A (2006) Modelling gap-size distribution of parked cars using
  random-matrix theory.
\newblock {\em Physica A: Statistical Mechanics and its Applications}
  368(2):536--540.

\bibitem{Jez10}
Jezbera D, Kordek D, K{\v{r}}{\'\i}{\v{z}} J, {\v{S}}eba P, {\v{S}}roll P
  (2010) Walkers on the circle.
\newblock {\em Journal of Statistical Mechanics: Theory and Experiment}
  2010(01):L01001.

\bibitem{Seb09}
{\v{S}}eba P (2009) Parking and the visual perception of space.
\newblock {\em Journal of Statistical Mechanics: Theory and Experiment}
  2009(10):L10002.

\bibitem{LecaDe93}
Le~Ca{\"e}r G, Delannay R (1993) The administrative divisions of mainland
  france as 2d random cellular structures.
\newblock {\em Journal de Physique I} 3(8):1777--1800.

\bibitem{LuZh06}
Luo F, Zhong J, Yang Y, Zhou J (2006) Application of random matrix theory to
  microarray data for discovering functional gene modules.
\newblock {\em Physical Review E} 73(3):031924.

\bibitem{Dei08}
Deift P (2008) Some open problems in random matrix theory and the theory of
  integrable systems.
\newblock {\em Contemporary Mathematics} 458:419.

\bibitem{Dei17}
Deift P (2017) Some open problems in random matrix theory and the theory of
  integrable systems. ii.
\newblock {\em arXiv preprint arXiv:1703.04931}.

\bibitem{W67}
Wigner EP (1967) Random matrices in physics.
\newblock {\em SIAM review} 9(1):1--23.

\bibitem{Ede13}
Edelman A, Wang Y (2013) Random matrix theory and its innovative applications
  in {\em Advances in Applied Mathematics, Modeling, and Computational
  Science}.
\newblock (Springer), pp. 91--116.

\bibitem{vNGo47}
Von~Neumann J, Goldstine HH (1947) Numerical inverting of matrices of high
  order.
\newblock {\em Bulletin of the American Mathematical Society}
  53(11):1021--1099.

\bibitem{DeiMen14}
Deift PA, Menon G, Olver S, Trogdon T (2014) Universality in numerical
  computations with random data.
\newblock {\em Proceedings of the National Academy of Sciences}
  111(42):14973--14978.

\bibitem{CarCer18}
Carmona R, Cerenzia M, Palmer AZ (2018) The dyson game.
\newblock {\em arXiv preprint arXiv:1808.02464}.

\bibitem{Led01}
Ledoux M (2001) {\em The concentration of measure phenomenon}, Mathematical
  Surveys and Monographs.
\newblock (American Mathematical Society, Providence, RI) Vol.{}~89.

\bibitem{Tsy09}
Tsybakov AB (2009) {\em Introduction to nonparametric estimation}, Springer
  Series in Statistics.
\newblock (Springer, New York), pp. xii+214.
\newblock Revised and extended from the 2004 French original, Translated by
  Vladimir Zaiats.

\bibitem{Wigner55}
Wigner EP (1955) Characteristic vectors of bordered matrices with infinite
  dimensions.
\newblock {\em Ann. of Math.} 62(3):548--564.

\bibitem{mehta2004random}
Mehta ML (2004) {\em Random matrices}.
\newblock (Elsevier) Vol.{} 142.

\bibitem{BaiBenPec05}
Baik J, Ben~Arous G, P\'{e}ch\'{e} S (2005) Phase transition of the largest
  eigenvalue for nonnull complex sample covariance matrices.
\newblock {\em Ann. Probab.} 33(5):1643--1697.

\bibitem{HouKriPer09}
Hough JB, Krishnapur M, Peres Y, Vir\'{a}g B (2009) {\em Zeros of {G}aussian
  analytic functions and determinantal point processes}, University Lecture
  Series.
\newblock (American Mathematical Society, Providence, RI) Vol.{}~51.

\bibitem{Kallenberg}
Kallenberg O (2006) {\em Foundations of modern probability}.
\newblock (Springer Science \& Business Media).

\bibitem{Bor11}
Borodin A (2011) Determinantal point processes in {\em The {O}xford handbook of
  random matrix theory}.
\newblock (Oxford Univ. Press, Oxford), pp. 231--249.

\bibitem{Joh01}
Johnstone IM (2001) On the distribution of the largest eigenvalue in principal
  components analysis.
\newblock {\em The Annals of Statistics} 29(2):295--327.

\bibitem{Pau07}
Paul D (2007) Asymptotics of sample eigenstructure for a large dimensional
  spiked covariance model.
\newblock {\em Statist. Sinica} 17(4):1617--1642.

\bibitem{FerPec09}
F{{\'e}}ral D, P{{\'e}}ch{{\'e}} S (2009) The largest eigenvalues of sample
  covariance matrices for a spiked population: diagonal case.
\newblock {\em J. Math. Phys.} 50(7):073302, 33.

\bibitem{And84}
Anderson TW (1984) {\em An introduction to multivariate statistical analysis},
  Wiley Series in Probability and Mathematical Statistics: Probability and
  Mathematical Statistics.
\newblock (John Wiley \& Sons, Inc., New York), Second edition.

\bibitem{BerRig13}
Berthet Q, Rigollet P (2013) Optimal detection of sparse principal components
  in high dimension.
\newblock {\em Ann. Statist.} 41(1):1780--1815.

\bibitem{BerRig13b}
Berthet Q, Rigollet P (2013) Complexity theoretic lower bounds for sparse
  principal component detection in {\em COLT 2013 - The 26th Conference on
  Learning Theory, Princeton, NJ, June 12-14, 2013}, JMLR W\&CP, eds.{}
  Shalev-Shwartz S, Steinwart I.
\newblock Vol.{}~30, pp. 1046--1066.

\bibitem{JolCad16}
Jolliffe IT, Cadima J (2016) Principal component analysis: a review and recent
  developments.
\newblock {\em Philos. Trans. Roy. Soc. A} 374(2065):20150202, 16.

\bibitem{BruMoiRig17}
Brunel VE, Moitra A, Rigollet P, Urschel J (2017) Rates of estimation for
  determinantal point processes in {\em Proceedings of the 2017 Conference on
  Learning Theory}, Proceedings of Machine Learning Research, eds.{} Kale S,
  Shamir O.
\newblock (PMLR), Vol.{}~65, pp. 343--345.

\bibitem{UrsBruMoi17}
Urschel J, Brunel VE, Moitra A, Rigollet P (2017) Learning determinantal point
  processes with moments and cycles in {\em Proceedings of the 34th
  International Conference on Machine Learning}, Proceedings of Machine
  Learning Research, eds.{} Precup D, Teh YW.
\newblock (PMLR), Vol.{}~70, pp. 3511--3520.

\bibitem{FIdata}
(1988) Fisher's iris data set (Available from the {UCI} Machine Learning
  Repository at \url{https://archive.ics.uci.edu/ml/datasets/Iris}).

\bibitem{Fis36}
Fisher RA (1936) The use of multiple measurements in taxonomic problems.
\newblock {\em Annals of eugenics} 7(2):179--188.

\bibitem{WBdata}
(1992) Wisconsin breast cancer data set (Available from the {UCI} Machine
  Learning Repository at
  \url{https://archive.ics.uci.edu/ml/datasets/Breast+Cancer+Wisconsin+(Original)}).

\bibitem{Wol90}
Wolberg WH, Mangasarian OL (1990) Multisurface method of pattern separation for
  medical diagnosis applied to breast cytology.
\newblock {\em Proceedings of the national academy of sciences}
  87(23):9193--9196.

\bibitem{Rig19}
Riogllet P (2019) R codes for data analyses (Available at
  \url{http://www-math.mit.edu/~rigollet/Code/DPP/DPPcode.R}).

\end{thebibliography}


\begin{thebibliography}{1}

\bibitem{LavMolRub15}
Lavancier F, M{\o}ller J, Rubak E (2015) Determinantal point process models and
  statistical inference.
\newblock {\em J. R. Stat. Soc. Ser. B. Stat. Methodol.} 77(4):853--877.

\bibitem{HouKriPer09}
Hough JB, Krishnapur M, Peres Y, Vir\'{a}g B (2009) {\em Zeros of {G}aussian
  analytic functions and determinantal point processes}, University Lecture
  Series.
\newblock (American Mathematical Society, Providence, RI) Vol.{}~51.

\bibitem{LauMas00}
Laurent B, Massart P (2000) Adaptive estimation of a quadratic functional by
  model selection.
\newblock {\em Ann. Statist.} 28(5):1302--1338.

\end{thebibliography}

\end{document}



\maketitle

\SItext

\section{Existence of Gaussian DPPs}
\label{sec:existence}
In this section, we show that Gaussian DPPs indeed exist. Note that existence in the isotropic case was already proved in~\cite{LavMolRub15} and we complete their result for all positive-definite scattering matrices.
\begin{theorem}\label{thm:existence}
For any positive-definite matrix $\Sigma$, there exists a stationary determinantal point process on $\R^d$ with the kernel $$
K(x,y)=\Phi(x-y)=\frac{1}{(2\pi)^{\frac{d}{2}} \sqrt{\det{\Sigma}}} \exp\left(-\frac{1}{2}(x-y)^\top \Sigma^{-1}(x-y)\right)
$$
and the Lebesgue measure on $\R^d$ as its background measure.
\end{theorem}

\begin{proof}
We may view $K(x,y)=\Phi(x-y)$ as the kernel of an integral operator $T$ acting on $L^2(\R^d)$, as
$$
Tf(x)=\int f(y)K(x,y)\ud y= f\star \Phi(x)\,,
$$
where $\star$ denotes convolution. Let us study at the spectrum of this integral operator. To that end, recall that Plancherel's theorem implies that the Fourier transform defined for any $f \in L^2(\R^d)$ by
$$
\widehat{f}(\omega):=\int_{\R^d} e^{-2\pi i \langle x, \omega \rangle }  f(x) \ud x
$$
is an isometry from $L^2(\R^d)$ to itself. Therefore, the spectrum of $T$ as an integral operator on $L^2(\R^d)$ is the same as the spectrum of the Fourier conjugate of this operator acting on $L^2(\R^d)$. Since $\widehat{\Phi \star f}=\widehat{\Phi}\cdot\widehat{f}$,  this Fourier conjugate is the operator of multiplication by $\widehat{\Phi}$. Since the spectrum of a multiplication operator is the (closure of the) range of the multiplier function,  our spectrum of interest is the (closure of the) range of the function $\widehat{\Phi}$. But $\widehat{\Phi}(\omega)=\exp(-2 \pi^2 \omega^\top \Sigma \omega )$, so the closure of its range, and hence our spectrum of interest, is $[0,1]$.

The above discussion shows that the integral operator $T$ is a non-negative contraction on $L^2(\R^d)$. Moreover, since $K(x,x)=\Phi(0)$ is independent of $x$, we clearly have $\int_B K(x,x) \ud x =\mathrm{Vol}(B) \cdot \Phi(0)<\infty$ for any bounded domain $B \subset \R^d$. In other words, the said integral operator is also locally trace class. By the celebrated Macchi-Soshnikov Theorem~\cite[Theorem~4.5.5]{HouKriPer09}, we deduce that there exists a DPP on $\R^d$ with kernel $K(x,y)=\Phi(x-y)$ and background measure the Lebesgue measure. Since the kernel depends only on the difference $x-y$, the resulting DPP is stationary.
\end{proof}

\begin{remark}
Notice that in our considerations in this paper, we have worked with Theroem \ref{thm:existence} under the normalization $(2\pi)^{\frac{d}{2}} \sqrt{\det{\Sigma}}=1$, which leads to the mean density of points being 1.
\end{remark}

\section{Concentration of the number of points $N$}

Let $X$ be a Gaussian DPP with scattering matrix $\Sigma$ and let $X_1,\ldots, X_{N} \in B(R)$ denote the set of points of $X$ that are included in the Euclidean ball $B(R)$ of radius $R>0$. Clearly, the number $N$ of these points is random.  The following theorem establishes sharp concentration of the random number of points $N$ around its expectation $n=\E[N]$.

\begin{theorem} 
The expectation $n$ of $N$ is given by
$$
n=\E[N]=|B(R)|=|B(1)|R^d\,,
$$
where $|B(t)|$ denotes the volume of Euclidean ball radius $t>0$ moreover, for 
\begin{equation}
	\p\big[\big| \frac{N}{n} - 1  \big| \ge \eps  \big]\leq 2\exp\big( - \frac{3\eps^2}{ 6 + 2\eps }|B(R)|\big).
\end{equation}
\end{theorem}
\begin{proof}
From the definition of DPPs, we can readily compute the expectation of $N$:
$$
n=\int_{B(R)}\rho_1(x) \ud x= \int_{B(R)}\det[K(x,x)] \ud x=|B(R)|=|B(1)|R^d\,.
$$
To study concentration, recall  \cite[Theorem~4.5.5]{HouKriPer09} that $N$ can be expressed as 
$$
N = \sum_{i=1}^{\infty} Z_i\,,
$$ 
where the $Z_i$s are independent Bernoulli random variables with parameter $\lambda_i$ respectively, where $\lambda_i$ is the $i$th eigenvalue (arranged in decreasing order) of the truncated integral operator $T_{|B(R)}$ obtained by restricting the integral operator $T$ defined in the previous section to the ball $B(R)$ of radius $R$.

This representation allows us to employ standard tools on the concentration of sums of independent random variables. By Bernstein's inequality, we get
\begin{align*}
\p(|N/\E[N] - 1|> \eps )\le2\exp\big( - \frac{\eps^2 n^2}{2\sum_{i=1}^\infty \E[Z_i^2] + 2 \eps n /3}  \big) 
&\le 2\exp\big( - \frac{3\eps^2 n}{6 + 2 \eps }  \big) =2\exp\big( - \frac{3\eps^2}{6 + 2 \eps}|B(R)|  \big),
\end{align*}
where in obtaining the second inequality we have used the fact that  $\sum_{i=1}^\infty \E[Z_i^2]  \le \sum_{i=1}^\infty \E[Z_i] = \E[N] = n$. It may noted that, strictly speaking, the Bernstein's inequality in its standard form considers concentration for sums of finitely many random variables, from which our desired statement about an infinite sum can be obtained by passing to the limit.
\end{proof}

\section{Estimation error}
In this section we prove our main statistical result. To avoid confusion with the Fourier transforms that appear in this supplementary information, we denote by $\tilde \Sigma$ (instead of $\hat \Sigma$ as in the main text), the estimator of $\Sigma$ defined by
$$
\tilde \Sigma = |B(1)|\frac{r^{d+2}}{d+2} I_d- \frac1{|B(R-r)|}\sum_{i \in \cN_0} \sum_{j \in \cN_i} (X_i-X_j)(X_i-X_j)^\top .
$$

\begin{theorem}
Assume that $\|\Sigma\|_{\mathrm{op}}^2$ is bounded above by a universal constant and let $r=C_0\sqrt{d\log n}$ for some sufficiently large universal constant $C_0$, Then 
\begin{equation*}
\E\|\tilde \Sigma -\Sigma\|_{\mathrm{F}} \lesssim d^2 \log R \left( \frac{ c \sqrt{d} \log R  }{R}\right)^{d/2},
\end{equation*}
\end{theorem}
We prove this theorem with a traditional bias-variance decomposition.

\subsection{Control of the bias}
In this section, we control the bias of our estimator.
The following lemma shows that our estimator is negatively biased but that its bias vanishes exponentially fast as $r \to \infty$. 

\begin{lemma}\label{lem:bias}
For any $v \in \R^d$ such that $\|v\|=1$, it holds for any $r\ge \sqrt{5 \tr(\Sigma)/2}$ that
\begin{equation}
\label{EQ:bound-bias01}
0 \le v^\top {\Sigma}v - \E[v^\top\tilde{\Sigma}v]  \le 3\lambda_1 \exp\big((2\tr(\Sigma)-r^2)/3\lambda_1 \big)\,.
\end{equation}
In particular, $\Sigma-\E\tilde \Sigma$ is positive-semidefinite and 
\begin{equation}
\label{EQ:bound-bias2}
\|\E\tilde \Sigma-\Sigma\|^2_{\mathrm{F}}\le 9d\|\Sigma\|_{\mathrm{op}}^2 \exp\big((4\tr(\Sigma)-2r^2)/3\|\Sigma\|_{\mathrm{op}}\big)\,.
\end{equation}
\end{lemma}
\begin{proof}
Observe that
$$
v^\top\tilde{\Sigma}v=|B(1)|\frac{r^{d+2}}{d+2}\|v\|^2 - \frac1{|B(R-r)|}\sum_{i \in \cN_0} \sum_{j \in \cN_i} \langle X_i-X_j, v\rangle^2\,.
$$

Recall that from the definition of the two-point correlation $\rho_2$, we have for any bounded function $\psi:\R^d\times \R^d \to \R$, that
$$
\E\sum_{i \ne j} \psi (X_i,X_j)=\iint_{B(R)\times B(R)} \psi(x,y) \rho_2(x,y) \ud x \ud y\,.
$$
Moreover, for Gaussian DPPs (recall our normalization $(2\pi)^{\frac{d}{2}} \sqrt{\det{\Sigma}}=1$), the two-point correlation $\rho_2$ is given by
$$
\rho_2(x,y)=\det \left[\begin{array}{cc}1 & \Phi(x-y) \\\Phi(x-y)   & 1\end{array}\right]=1-\exp\big(-(x-y)^\top \Sigma^{-1} (x-y)\big)\,.
$$
Therefore, choosing the test function
$$
\psi(x,y)=-\frac1{|B(R-r)|}\langle x-y, v\rangle^2 \1\{\|x-y\|\le r\}\1\{\|x\| \le R-r\}\,,
$$
we get
\begin{align*}
\E[v^\top \tilde{\Sigma} v]&=|B(1)|\frac{r^{d+2}}{d+2}\|v\|^2-\E\sum_{ i \ne j} \psi (X_i,X_j)\\
&=|B(1)|\frac{r^{d+2}}{d+2}-\frac1{|B(R-r)|}\iint_{\substack{\|x-y\|\le r\\\|x\| \le R-r}}\langle x-y, v\rangle^2\big(1-e^{-(x-y)^\top \Sigma^{-1} (x-y)}  \big)\ud x \ud y \numberthis \label{EQ:explicit-integral}
\end{align*}
For a fixed $x$, we first compute the integral with respect to $\ud y$ given by
$$
\int_{\|x-y\|\le r}\langle x-y, v\rangle^2\big(1-e^{-(x-y)^\top \Sigma^{-1} (x-y)}  \big) \ud y\,.
$$
Using the change of variables $u=x-y$ it is equal to
\begin{align*}
\int_{\|u\|\le r}\langle u, v\rangle^2\big(1-e^{-u^\top \Sigma^{-1} u}  \big) \ud u&=\int_{\|u\|\le r}\langle u, v\rangle^2\ud u-\int_{\|u\|\le r}\langle u, v\rangle^2 e^{-u^\top \Sigma^{-1} u}  \ud u\\
&=r^2|B(r)|\E[\langle U, v\rangle^2]-\E[\langle  \xi, v\rangle^2\1\{\|\xi\|\le r\}]   \numberthis \label{EQ:cov}  \\ 
\end{align*}
where $U\sim \mathsf{Unif}(B(1))$ and $\xi\sim \mathsf{N}(0,\Sigma)$ and we used the normalization $\sqrt{(2\pi)^{d}\det(\Sigma)}$=1. Next, note that by elementary computations and the fact that $\|v\|^2=1$, we have
$$
\E[\langle U, v\rangle^2]=v^\top \E[UU^\top] v=\frac{1}{d+2} v^\top I_d v=\frac{1}{d+2}\,,
$$
so that
\begin{equation} \label{EQ:inner-integral}
\int_{\|u\|\le r}\langle u, v\rangle^2\big(1-e^{-u^\top \Sigma^{-1} u}  \big) \ud u=\frac{|B(1)|r^{d+2}}{d+2}-\E[\langle  \xi, v\rangle^2\1\{\|\xi\|\le r\}]\,.
\end{equation}
Note that the above expression does not depend on $x$, so combining \eqref{EQ:explicit-integral}, \eqref{EQ:cov} and \eqref{EQ:inner-integral}, we obtain
$$
\E[v^\top \tilde{\Sigma} v]=\E[\langle  \xi, v\rangle^2\1\{\|\xi\|\le r\}]\to \E[\langle  \xi, v\rangle^2]=v^\top \Sigma v\,, \quad \text{as }r \to \infty
$$
Moreover, $\E[v^\top \tilde{\Sigma} v] \le v^\top \Sigma v$.
To control the rate at which this convergence takes place, note that by the Cauchy-Schwarz and Fubini theorems, we get
\begin{align}
\E[\langle  \xi, v\rangle^2\1\{\|\xi\|\le r\}]-\E[\langle  \xi, v\rangle^2]
&=\E[\langle  \xi, v\rangle^2\1\{\|\xi\|> r\}\le \E[\| \xi\|^2\1\{\|\xi\|> r\}]
 = \int_{r^2}^\infty\p\big(\| \xi\|^2>t\big) \ud t\,.\label{EQ:inttails}
\end{align}

Let $\lambda=(\lambda_1, \ldots, \lambda_d)^\top$ denote the vector of eigenvalues of $\Sigma$ ordered in a nonincreasing fashion: $\lambda_1\ge \lambda_2 \ge \dots\ge \lambda_d>0$. Observe that $\| \xi\|^2$ has the same distribution as $\lambda_1g_1^2+\cdots+ \lambda_dg_d^2$,
where $g_1, \ldots, g_d$ are i.i.d standard Gaussian random variables. Thus, it follows from~\cite[Lemma~1]{LauMas00} that for any $x$, we have
$$
\p(\| \xi\|^2 \ge \tr(\Sigma)+ 2\|\lambda\|x+2\lambda_1x^2)\le \exp(-x^2)\
$$
For $t \ge \tr(\Sigma)$, the equation  
\begin{equation} \label{EQ:quadratic} t=\tr(\Sigma)+ 2\|\lambda\|x+2\lambda_1x^2 \end{equation} 
 has a unique solution in $x$ given by
$$
x=\frac{\sqrt{4\|\lambda\|^2-8\lambda_1\big(\tr(\Sigma)-t\big)}-2\|\lambda\|}{4\lambda_1}\,.
$$
Using the fact that $\|\lambda\|^2\le \lambda_1 \tr(\Sigma)$, we obtain $x^2 \ge (t- 2\tr(\Sigma))/3\lambda_1$ for $t \ge 5 \tr(\Sigma)/2$. Indeed, we have 
$$ 2\|\lambda\|x \le 2 \sqrt{\tr(\Sigma)} \cdot \sqrt{\lambda_1}x  \le  \tr(\Sigma) +  \lambda_1 x^2 ,$$ and applying this to \eqref{EQ:quadratic} we deduce that  $t \le 2\tr(\Sigma)+ 3\lambda_1x^2$. Rearranging terms, we deduce that $x^2 \ge (t- 2\tr(\Sigma))/3\lambda_1$, as desired.
 It yields that for $r^2\ge 5 \tr(\Sigma)/2$,
$$
\int_{r^2}^\infty\p\big(\| \xi\|^2>t\big) \ud t \le 3\lambda_1 \exp\big((2\tr(\Sigma)-r^2)/3\lambda_1 \big)\,.
$$
Plugging this result into~\eqref{EQ:inttails} concludes the proof of \eqref{EQ:bound-bias01}. 

To obtain the bound \eqref{EQ:bound-bias2} on the Frobenius norm, observe that
$$\|\E\tilde \Sigma-\Sigma\|^2_{\mathrm{F}}\le d\|\E\tilde \Sigma-\Sigma\|^2_{\mathrm{op}}=d\sup_{v\,:\|v\|=1}[v^\top(\Sigma-\E\tilde \Sigma)v]^2 \le  9d\|\Sigma\|_{\mathrm{op}}^2 \exp\big((4\tr(\Sigma)-2r^2)/3\|\Sigma\|_{\mathrm{op}}\big)\,.$$
\end{proof}


\subsection{Control of the variance}
The variance term is controlled by the following lemma.
\begin{lemma}\label{lem:var}
Let $u,v \in \R^d$ be two vectors such that $\|u\|=\|v\|=1$. It holds for any $r$ such that $\sqrt{d}\le r \le R/2$ that
\begin{equation} \label{EQ:lemvar1}
\var(u^\top\tilde \Sigma v)\le\left(\frac{C}{d}\right)^d\frac{r^{2d+4}}{n}
%
\end{equation}
for a universal constant $C>0$. In particular,
\begin{equation} \label{EQ:lemvar2}
\E\|\hat{\Sigma} - \E\hat{\Sigma}\|_{\mathrm{F}}^2\le d^2\left(\frac{C}{d}\right)^d\frac{r^{2d+4}}{n}
\end{equation}
\end{lemma}
\begin{proof}


We have
\begin{align*}
u^\top\tilde \Sigma v&=|B(1)|\frac{r^{d+2}}{d+2}\langle u,v\rangle-\frac1{|B(R-r)|}\sum_{i \in \cN_0} \sum_{j \in \cN_i} \langle X_i-X_j, u\rangle \cdot\langle X_i-X_j, v\rangle\\
&=|B(1)|\frac{r^{d+2}}{d+2}\langle u,v\rangle-\frac1{|B(R-r)|}\sum_{i\neq j}  \langle X_i-X_j, u\rangle \cdot\langle X_i-X_j, v\rangle\1\{\|X_i-X_j\|\le r, \|X_i\|\le R-r\}\,.
\end{align*}
It yields
\begin{equation}
\label{EQ:boundvar0}
\var(u^\top\tilde \Sigma v)\le \frac{1}{|B(R-r)|^2}\var\left(\sum_{i \neq j}\psi(X_i,X_j)\right)\,.
\end{equation}
where
$$
\psi(x,y)=\langle x-y, u\rangle\langle x-y, v\rangle\1\{\|x-y\|\le r\}\1\{\|x\|\le R-r\}\,.
$$
 
It holds
\begin{align*}
&\var\left(\sum_{i \neq j}\psi(X_i,X_j)\right) \\= &\sum_{\substack{i \neq j\\k\neq l}}\E[\psi(X_i,X_j)\psi(X_k,X_l)]-\left(\E\sum_{ i \neq j }\psi(X_i,X_j)\right)^2\\
=&\iiiint \psi(x,y)\psi(z,w)\rho_4(x,y,z,w)\ud x\ud y\ud z\ud w \\
&\quad + \iiint \psi(x,y)\psi(y,w)\rho_3(x,y,w)\ud x\ud y\ud w  + \iiint \psi(x,y)\psi(x,w)\rho_3(x,y,w)\ud x\ud y\ud w \\
&\quad + \iiint \psi(x,y)\psi(z,y)\rho_3(x,y,z)\ud x\ud y\ud z  + \iiint \psi(x,y)\psi(z,x)\rho_3(x,y,z)\ud x\ud y\ud z \\
&\qquad + \iint \psi(x,y)^2\rho_2(x,y)\ud x\ud y - \left(\iint \psi(x,y) \rho_2(x,y)\ud x \ud y\right)^2\,.
\end{align*}
By symmetry of $K$ (and thus of $\rho_3$), all triple integrals are the same, equal to
$$
\iiint \psi(x,y)\psi(x,z)\rho_3(x,y,z)\ud x\ud y\ud z\,.
$$
Moreover, since $4$-point correlation $\rho_4$ is given by
$$
\rho_4(x_1, x_2,x_3,x_4)=\det \left[ \begin{array}{cccc}
K(x_1,x_1)  & K(x_1,x_2) &  K(x_1,x_3) & K(x_1,x_4) \\
K(x_2,x_1)  & K(x_2,x_2) &  K(x_2,x_3) & K(x_2,x_4) \\
K(x_3,x_1)  & K(x_3,x_2) &  K(x_3,x_3) & K(x_3,x_4) \\
K(x_4,x_1)  & K(x_4,x_2) &  K(x_4,x_3) & K(x_4,x_4)
\end{array}\right]\,,
$$
it follows from the symmetry of $K$ and Fischer's inequality that
\begin{equation}
\label{EQ:rho4}
\rho_4(x_1, x_2,x_3,x_4)\le \det \left[ \begin{array}{cc}
K(x_1,x_1)  & K(x_1,x_2) \\
K(x_2,x_1)  & K(x_2,x_2)
\end{array}\right]\cdot \det \left[ \begin{array}{cc}
K(x_3,x_3)  & K(x_3,x_4) \\
K(x_4,x_3)  & K(x_4,x_4)
\end{array}\right]=\rho_2(x_1,x_2)\rho_2(x_3,x_4)\,.
\end{equation}
Therefore,
$$
\iiiint \psi(x,y)\psi(z,w)\rho_4(x,y,z,w)\ud x\ud y\ud z\ud w -\left(\iint \psi(x,y) \rho_2(x,y)\ud x \ud y\right)^2\le 0
$$
Together, these facts yield
\begin{equation}
\label{EQ:boundvar}
\var\left(\sum_{i \neq j}\psi(X_i,X_j)\right)\le 4\iiint \psi(x,y)\psi(x,z)\rho_3(x,y,z)\ud x\ud y\ud z+\iint \psi(x,y)^2\rho_2(x,y)\ud x\ud y\,.
\end{equation}

We control both terms in the right-hand side above in a similar fashion. On the one hand, from Fischer's inequality, it holds that for any positive integer $r$,
\begin{equation}
\label{EQ:rhok}
\rho_r(x_1, \dots, x_r)\le \prod_{i=1}^rK(x_i,x_i)=1\,.
\end{equation}
On the other hand, from the Cauchy-Schwarz inequality, we have
$$
\psi(x,y)\le r^2\1\{\|x-y\|\le r\}\1(\|x\|\le R-r\}\,.
$$
Together, the above two displays yield
\begin{align*}
\iiint \psi(x,y)\psi(x,z)\rho_3(x,y,z)\ud x\ud y\ud z&\le r^4\iiint\1\{\|x-y\|\le r\}\1\{\|x-z\|\le r\}\1\{\|x\|\le R\}\ud x\ud y\ud z\\
&= r^4\int_{B(R-r)}\left( \int \1\{\|x-y\|\le r\} \ud y\right)^2 \ud x\\
&=r^4|B(r)|^2|B(R-r)|\\
&=r^{2d+4}(R-r)^d|B(1)|^3\,.
\end{align*}

Similarly,
\begin{align*}
\iint \psi(x,y)^2\rho_2(x,y)\ud x\ud z&\le r^4\iint\1\{\|x-y\|\le r\}\1\{\|x\|\le R-r\}\ud x\ud y=r^{d+4}(R-r)^d|B(1)|^2\,.
\end{align*}

Together with~\eqref{EQ:boundvar}, the above two displays establish that, for $r\ge \sqrt{d} \ge |B(1)|^{-1/d}$, we have
$$
\var\left(\sum_{i \neq j}\psi(X_i,X_j)\right)\le 5r^{2d+4}(R-r)^d|B(1)|^3
$$
Plugging this bound into~\eqref{EQ:boundvar0}, we get
$$
\var(u^\top\tilde \Sigma v)\le \frac{5}{(R-r)^d}r^{2d+4}|B(1)| \le \frac{5\cdot 2^d}{R^d}r^{2d+4}|B(1)|=\frac{5\cdot 2^d}{n}|B(1)|^2r^{2d+4}\,.
$$

Finally, using   Stirling's formula, we get $|B(1)|^{1/d} \le 2\pi e/\sqrt{d}$,
which concludes the proof of the first statement, i.e. \eqref{EQ:lemvar1}.

The second statement, i.e. \eqref{EQ:lemvar2}, follows by taking $u=e_i, v=e_j$, two vectors in the canonical basis of $\R^d$ and summing over the indices $1\le i,j\le d$. 
\end{proof}

\subsection{Bias-variance tradeoff}
We are now in a position to choose $r$ to realize the optimal bias-variance tradeoff.

Observe that by the AM-GM inequality
$$
\tr(\Sigma)\ge d [\det(\Sigma)]^{1/d}=\frac{d}{2\pi}
$$
From Lemmas~\ref{lem:bias} and~\ref{lem:var}, we get for $r \ge \sqrt{2\pi\tr(\Sigma)}$,
\begin{align*}
\E\|\tilde \Sigma -\Sigma\|_{\mathrm{F}}^2&=\|\E\tilde \Sigma -\Sigma\|_{\mathrm{F}}^2+\E\|\tilde \Sigma -\E\tilde\Sigma\|_{\mathrm{F}}^2\\
&\le 9d\|\Sigma\|_{\mathrm{op}}^2 \exp\big((4\tr(\Sigma)-2r^2)/3\|\Sigma\|_{\mathrm{op}}\big)+  d^2\left(\frac{C}{d}\right)^d\frac{r^{2d+4}}{n}\,.
\end{align*}

Assuming that $\|\Sigma\|_{\mathrm{op}}^2$ is bounded above by a universal constant, we get that the choice $r=C_0\sqrt{d\log n}$ for some sufficiently large universal constant $C_0$ yields
$$
\E\|\tilde \Sigma -\Sigma\|_{\mathrm{F}}^2 \le \frac{d^4(C \log n)^{d+2}}{n}\,.
$$
whenever $R>2r$.

\bibliography{dpp}